\def\year{2020}\relax
\documentclass[letterpaper]{article} 
\usepackage{aaai20}  
\usepackage{times}  
\usepackage{helvet} 
\usepackage{courier}  
\usepackage[hyphens]{url}  
\usepackage{graphicx} 
\usepackage{color}  
\usepackage{tablefootnote}
\usepackage{graphicx}  
\title{Retrieval-Augmented Transformer-XL for \\ Close-Domain Dialog Generation}
\author{\\ \Large \textbf{Giovanni Bonetta,\textsuperscript{\rm 1}\textsuperscript{\rm 2} Rossella Cancelliere,\textsuperscript{\rm 1} Ding Liu,\textsuperscript{\rm 2} Paul Vozila\textsuperscript{\rm 2}}  \\ 
\textsuperscript{\rm 1}Department of Computer Science, University of Turin, Torino, Italy\\
\textsuperscript{\rm 2}Nuance Communications Inc., Burlington, MA, USA \\
\{giovanni.bonetta, rossella.cancelliere\}@unito.it, \{ding.liu, paul.vozila\}@nuance.com\\
}
 
\begin{document}

\maketitle

\begin{abstract}
Transformer-based models have demonstrated excellent capabilities of capturing patterns and structures in natural language generation and achieved state-of-the-art results in many tasks.
In this paper we present a transformer-based model for multi-turn dialog response generation. Our solution is based on a hybrid approach which augments a transformer-based generative model with a novel retrieval mechanism, which leverages the memorized information in the training data via k-Nearest Neighbor search. 
Our system is evaluated on two datasets made by customer/assistant dialogs: the Taskmaster-1, released by Google and holding high quality, goal-oriented conversational data and a proprietary dataset collected from a real customer service call center. Both achieve better BLEU scores over strong baselines.
\end{abstract}

\section{Introduction}
\label{sec:introduction}
Automatic dialog generation is become today a fundamental component for many real-world, challenging applications, such as virtual assistants, chatbots, etc., and is also a matter of great concern for companies and organizations relying on artificial intelligence solutions to enhance millions of daily interactions through their services. 

Simple single-turn Seq2Seq architectures, initially proposed for this task, often fail to capture long-term temporal dependencies across dialog turns. \cite{sutskever2014sequence,Vinyals2015,Li-etal-2016-diversity}.
Multi-turn Seq2Seq models, such as the hierarchical recurrent encoder decoder (HRED) \cite{AAAI1611957,AAAIXing,serban2017b} have tried to alleviate these problems, yielding responses more coherent with the dialog contexts.
Nonetheless, the generated texts tend to be either generic or too short, and not comparable with the human ones.
Recently, pretrained transformer-based models such as BERT \cite{Devlin2018}, Transformer-XL \cite{Dai2019}, XLNet \cite{Yang2019} and ERNIE \cite{Zhang2019} led to state-of-the-art performance on many natural language processing/understanding (NLP/NLU) tasks, including question answering, sentence classification, sentence similarity inference, and named entity recognition etc.

An interesting idea which further enhances the generative model performance is to condition the generation on samples retrieved from a task-related datastore.
 In \cite{Kelvin2020,Kenton2019} a generative model is augmented with a neural retriever trained to pick informative text paragraphs; \cite{Khandelwal2020} propose to enhance a language model (LM) through a nearest neighbor search in suitable text collections. The model we present in this paper exploits a similar framework for dialog generation. Our first original contribution is showing how to generate dialog continuations using a LM augmented with a k-nearest neighbors (kNN) based retrieval mechanism.
Furthermore, we exploit the typical dialog structure to enhance and speed the retrieval mechanism, improving the generation results.
In section "Model Overview" we introduce our model and formally define our approach, also going into detail of the retrieval mechanism. The remaining sections are devoted to the dataset descriptions and results discussion.

\section{Model Overview}
\label{sec:Ourapproach}

We propose a method which improves dialog generation by exploiting memorized information from the training data, without further model training. 
At inference, turn generation is enhanced by interpolating the next word distribution based on the trained LM with the one based on a kNN search system. A single LM forward pass over the training data is preliminary conducted to compute context-target pairs and store them in a key-value pair {\it datastore}, which will be queried to perform the kNN search. The next sections describe this procedure and how a kNN distribution is computed and used to augment the LM.

\subsection{Datastore Creation}
\label{sec:System}

The first step in order to create the datastore is the training of a LM, in our case a Transformer-XL \cite{Dai2019}, by minimizing the cross entropy of the training data. Overfitting is controlled  through early stopping on validation data performance. 
Differently from \cite{Dai2019} and \cite{Khandelwal2020}, which train a LM by concatenating all the examples, we train the model by resetting the Transformer-XL states at the beginning of each chat: this effectively prevents the model from conditioning on previous unrelated contexts. 

Let $(c_t^i, w_t^i) \in D$ be the $i^{th}$ example in training data $D$. The context $c_t^i$ is a sequence of dialog turns of a dyadic chat occurring between an assistant and a user; $c_t^i$ is represented as a sequence of tokens, i.e. $c_t^i = (w_1^i, w_2^i \dots w_{t-1}^i)$, and $w_t^i$ is the target word.

Let $f({c^i_t})$ denote the context-encoder function, that maps the context ${c^i_t}$ to its fixed-length vector embedding. 
We define $f(\cdot)$ as the input to the last feedforward layer in the final attention block of Transformer-XL, as in \cite{Khandelwal2020}. This achieves better performance than other options (e.g, the output of the last transformer layer). More specifically, $f(c_t^i)$ represents the embedding of token $w_{t-1}^i$ after attending to all the previous tokens in the example. 

Through one forward pass on the training data, the trained LM is used to build the datastore $(K,W)$ containing the embeddings of all the tokens in the training data: 
$$
(K,W)\!:=\!(k_t^i,w_t^i)\!=\!(f(c_t^i),w_t^i),\:\: \:\:\forall(c_t^i, w_t^i)\!\in\!D
$$
where $k_t^i=f(c_t^i)$ is the vector representation of the context, and $w_t^i$ is the target word id (i.e. integer number). 
\subsection{Hybrid Probability Distribution}
\label{sec:Generative}

\begin{figure}
    \includegraphics[width=0.5\textwidth]{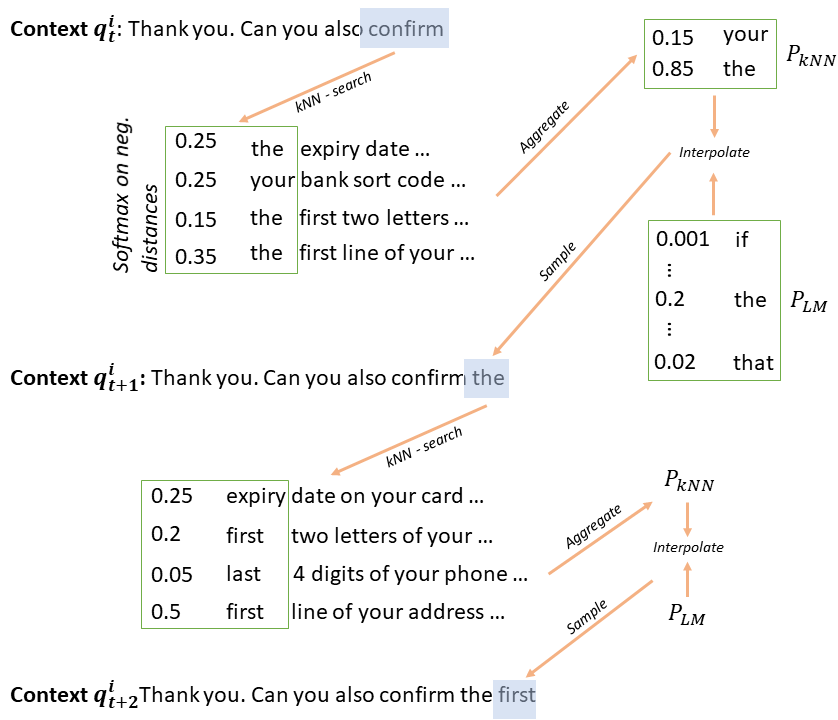}
    \caption{Illustration of the Generation Process}
    \label{fig:system}
\end{figure}
At inference, at every time step $t$, the trained LM receives a {\it query} ($q_t$), i.e. a chat truncated at the end of a user turn, and generates the next assistant turn token-by-token, according to the following steps, also illustrated in Fig.~\ref{fig:system}:
\begin{itemize}
    \item Generate the context embedding $f(q_t)$ and the probability distribution $P_{LM}(v_t|q_t)$ over next words in the vocabulary
    
    \item Issue a kNN search with $f(q_t)$ as query, to get from the datastore its nearest neighbors $N_t$:
$$
N_t = \{(k_1, w_1), (k_2, w_2) \dots (k_n, w_n)\dots\}
$$
      \item Compute the score $S_{kNN}(w_n|q_t)$ of the token $w_n$ over $N_t$, based on $L^2$ distance between $k_n$ and $f(q_t)$:  
$$
S_{kNN}(w_n|q_t) = \frac{e^{-d (k_n, f(q_t))} }{\sum_{k_j \in N_t} e^{-d (k_j, f(q_t))}} 
$$

    \item Aggregate the scores of each vocabulary token $ w_n $ as the sum of all its occurrences within the retrieved neighbors: 
$$
    S_{kNN}^{Aggr}(w_n|q_t) = \sum_{w_{n'} \in N_t \\ w_{n'} = w_{n}} S_{kNN}(w_{n'}|q_t)
$$

    \item Get the probability distribution $P_{kNN}$ over next words in the vocabulary:
    $$
    P_{kNN}(v_t|q_t)\! =\!\!\!\!\!\! \sum_{(k_n,w_n) \in N_t}\!\!\! \mathbf{1}_{v_t=w_n}  (S_{kNN}^{Aggr}(w_n|q_t))
    $$
 where $\mathbf{1}_{v_t=w_n}$ is a vector whose dimension is equal to the vocabulary size and whose elements are all zero except for the t-th one, equal to 1.

    \item Interpolate $P_{kNN}$ with $P_{LM}$ to get the final probability distribution $ P $ for next word $ v_t$ : 
    $$
    P(v_t|q_t)\! =\! \lambda P_{kNN}(v_t|q_t)\! +\! (1\!\! -\!\! \lambda) P_{LM}(v_t|q_t) 
    $$

    \item Sample the next word $\hat{v}_t$ by greedily sampling from $P(v_t|q_t)$ and concatenate $\hat{v}_t$ to $q_t$ to update the context: $q_{t+1} = q_t + \hat{v}_t$ 
    
\end{itemize}
    
\noindent If $\hat{v}_t$ is a terminal token the generation process stops; otherwise the entire procedure is repeated. 


\subsection{Retrieval Mechanism}
\label{sec:Retrieval}

To search the datastore, we use FAISS ~\cite{Johnson2017}, an open source library for fast nearest neighbor retrieval in high dimensional space. FAISS's central building block is the {\it index}, a structure which stores millions of key-value pairs for efficient search.
An issue with the index is that the number of elements could easily grow to hundreds of millions, leading to memory issues and hindering the search performance. 
However in practice, we only need to store token embeddings for assistant turns, since we are only interested in generating assistant responses.
So we propose the simple but effective idea of filtering out from the datastore every token coming from a user turn, so almost halving its size, and allows the generation of consistent utterances, resembling assitant specific style.


\section{Dataset Description}
\label{sec:datasets}
Two different datasets are used as benchmarks for our method: a public dataset, the Taskmaster-1, released by Google in 2019 and a real, company collected, call center customer service dataset.

\noindent \textbf{Taskmaster-1 dataset}. 
Taskmaster-1 \cite{byrne2019} is a crowsurced dataset, where Amazon turkers were asked to write dyadic dialogs following some given set of instructions describing six tasks: ordering pizza, creating auto repair appointments, setting up rides for hire, ordering movie tickets, ordering coffee drinks and making restaurant reservations. Workers were asked to play the role of both assistant and user. Specifically, they were told to write a scenario in
which they are speaking to their assistant on the
phone while the assistant accesses the services for
one of the given tasks. 
The resulting dataset contains 7,708 conversations.
More info about the dataset are in table \ref{tab:data_stats}.

\noindent \textbf{Proprietary (Prop.) dataset}.\footnote{The dataset can not be made public due to privacy constraints} This dataset contains dyadic agent-user chats collected from a financial service call center over a one year time period, giving us the opportunity to test our approach in a real company scenario. It contains 172 times the dialogs number of the Taskmaster-1, as shown in table \ref{tab:data_stats}, and comes with two meta-information, the turn numbers and the agent-ids. The turn number is just the position of the specific turn within the chat, while the agent-id is a unique identifier for the agent speaking. We concatenate these information to the chat's text, following the approach used in \cite{Wolf2019}. An example is given in figure \ref{fig:sample}.

\begin{table}
\caption{Dataset specifications.}
\vspace{1em}
\label{tab:data_stats}
\begin{tabular}{lll}
                     & \textbf{Taskmaster-1} & \textbf{Prop. dataset} \\ \hline
\# dialogs           & 7,708                 & 1,328,301              \\
\# turns             & 169,467               & 21,953,321             \\
\# unique tokens     & 29,626                & 1,601,647              \\
avg. turn per chat   & 21.99                 & 16.53                  \\
avg. tokens per turn & 7.83                  & 18.00                 
\end{tabular}
\end{table}



\section{Implementation Details and Results}
\label{sec:results}
In this section we present the model implementation details and discuss the results obtained for both datasets.

\subsection{Taskmaster-1 dataset}
For the Taskmaster-1 we used a Transformer-XL model with 12 layers, 8 heads, 512-dimensional  hidden states and 2048 as inner attention dimension, resulting in 49M weights and trained for a maximum of 10k steps optimizing with Adam. The training stopping criterion is based on perplexity on the development set. Hyperparameter tuning, including optimal $\lambda$ determination, is done through performance evaluation over the development set.
We adopted a BPE vocabulary (Sennrich, Haddow, and Birch2015) consisting of 16K tokens and generated using the Sentencepiece library (Kudo and Richardson 2018).
All the training set is used to build the datastore.

Our model Transformer-XL + kNN is compared with two baselines:  -Transformer, the best performing model by \cite{byrne2019} and - Transformer-XL, i.e. the LM used without the retrieval mechanism.
The first column of table~\ref{BLEU} shows the corresponding BLEU scores\footnote{BLEU script at: https://github.com/tensorflow/tensor2tensor/blob\\/master/tensor2tensor/bin/t2t-bleu}, obtained as mean values of 10 different runs, and standard deviations. 
We can see that our method gets more than two BLEU points over the Transformer baseline, and more than one point over the Transformer-XL baseline. 

\begin{table}[htbp]
\centering
\caption{Average BLEU and standard deviations on test set. The statistical significance is validated via Student's $t$-test with significance level of $99.8\%$.}
\label{BLEU} 
\vspace{1em}
\begin{tabular}{lcccc}
\multicolumn{1}{c}{} & \multicolumn{2}{c}{\textbf{Taskmaster-1}} & \multicolumn{2}{c}{\textbf{Prop. dataset}} \\
\hline
\textbf{Models}: & Avg & Std & Avg & Std \\
\hline
Transf.  & 6.11\tablefootnote{Results from original paper} & - & - & - \\
\cite{byrne2019}  & & & \\
Transf.-XL & 7.09 & 0.14 & 39.96 &  0.36\\
Transf.-XL + kNN & \textbf{8.30} & 0.05 & \textbf{41.72} &  0.20\\
\end{tabular}
\label{tab:cnn-mlp}
\end{table}

\begin{figure}
    \includegraphics[width=0.40\textwidth]{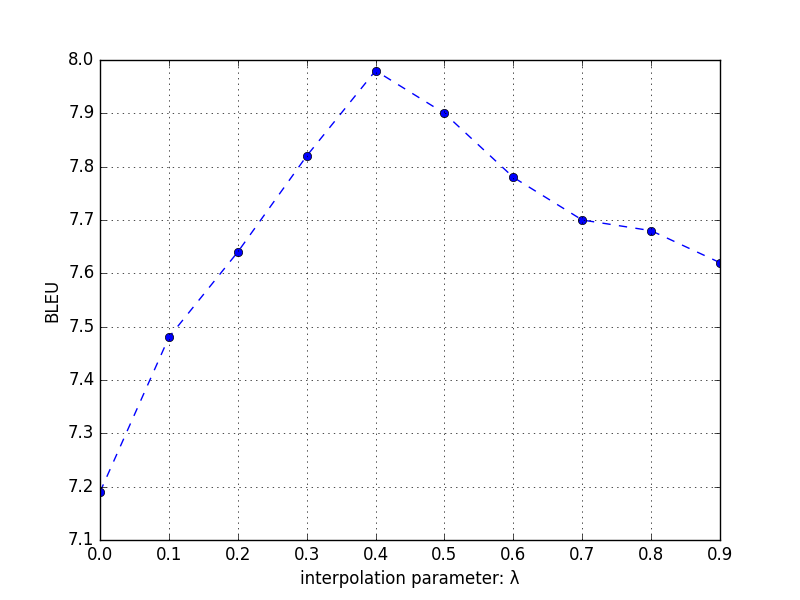}
    \caption{Taskmaster-1 BLEU trend (development set).}
    \label{fig:dev_results}
\end{figure}

Figure~\ref{fig:dev_results} depicts the BLEU trend curve when the interpolation parameter $\lambda$ varies through the selected range. We can see that kNN interpolation improves the BLEU scores over the Transformer-XL baseline for every value of $\lambda$ in the selected range. The best result is with $\lambda=0.4$, indicating LM and context retrieval are almost equally contributing.

\subsection{Proprietary dataset}
For the proprietary dataset we used the same model hyperparameters as for the Taskmaster-1 but augmented the hidden states dimension to 768 and the inner attention dimension to 3072, resulting in \textasciitilde116M weights. We trained for a maximum of 400k steps.

Since using all the training set for the datastore would result in a prohibitively large disk space usage we decided to build it using just the last 3 months of the training set ($1/4$ of the entire data). This resulted in \textasciitilde176M embeddings which occupy \textasciitilde500GB of disc memory. 
Also in this case the Transformer-XL + kNN improves over the LM model for about 1.8 BLEU points, even with a datastore smaller then the entire training set. These results are obtained interpolating with $\lambda = 0.5$ (best on dev. set).

\begin{figure}
    \includegraphics[width=0.5\textwidth]{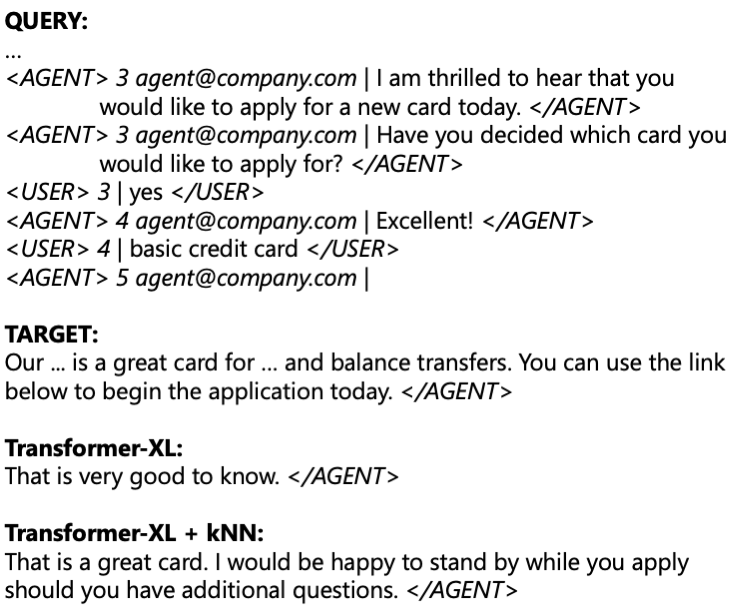}
    \caption{Example of inference query, along with results from baseline and our best model. \textit{agent@company.com} is the agent-id, which is preceded by the turn number. Tokens between angular parenthesis indicate the beginning and end of turns. }
    \label{fig:sample}
\end{figure}

Figure~\ref{fig:sample} shows a sample from the test data along with the expected target, the turn generated by the Transformer-XL, and the turn generated by our Transformer + kNN. In this dialog a user wants some help for a credit card application.
Our proposed model generates a sensible and relevant continuation: the agent conveys the intent to help the user apply for the credit card, as in the target. On the other hand the baseline Transformer-XL model generates a generic response which is not useful in advancing the dialog.

\section{Conclusions}
In this work we shown how to enhance a generative model for dialog completion by pairing it with an effective retrieval system. Our approach achieves higher BLEU scores than strong generative models when tested on two challenging datasets. Moreover, our solution often outputs more sensible/informative dialog turns.
In the future we plan to extend this preliminary work analysing more models on different datasets, and further investigating results and generated examples.

\bibliographystyle{aaai}
\bibliography{flairs34_paper}

\begin{thebibliography}{}

\bibitem[\protect\citeauthoryear{Byrne \bgroup et al\mbox.\egroup
  }{2019}]{byrne2019}
Byrne, B.; Krishnamoorthi, K.; Sankar, C.; Neelakantan, A.; Goodrich, B.;
  Duckworth, D.; Yavuz, S.; Dubey, A.; Kim, K.; and Cedilnik, A.
\newblock 2019.
\newblock Taskmaster-1: Toward a realistic and diverse dialog dataset.
\newblock In Inui, K.; Jiang, J.; Ng, V.; and Wan, X., eds., {\em Proceedings
  of the 2019 Conference on Empirical Methods in Natural Language Processing
  and the 9th International Joint Conference on Natural Language Processing,
  {EMNLP-IJCNLP} 2019, Hong Kong, China, November 3-7, 2019},  4515--4524.
\newblock Association for Computational Linguistics.

\bibitem[\protect\citeauthoryear{Dai \bgroup et al\mbox.\egroup
  }{2019}]{Dai2019}
Dai, Z.; Yang, Z.; Yang, Y.; Carbonell, J.~G.; Le, Q.~V.; and Salakhutdinov, R.
\newblock 2019.
\newblock Transformer-xl: Attentive language models beyond a fixed-length
  context.
\newblock {\em CoRR} abs/1901.02860.

\bibitem[\protect\citeauthoryear{Devlin \bgroup et al\mbox.\egroup
  }{2018}]{Devlin2018}
Devlin, J.; Chang, M.; Lee, K.; and Toutanova, K.
\newblock 2018.
\newblock {BERT:} pre-training of deep bidirectional transformers for language
  understanding.
\newblock {\em CoRR} abs/1810.04805.

\bibitem[\protect\citeauthoryear{Guu \bgroup et al\mbox.\egroup
  }{2020}]{Kelvin2020}
Guu, K.; Lee, K.; Tung, Z.; Pasupat, P.; and Chang, M.
\newblock 2020.
\newblock {REALM:} retrieval-augmented language model pre-training.
\newblock {\em CoRR} abs/2002.08909.

\bibitem[\protect\citeauthoryear{Johnson, Douze, and
  J{\'e}gou}{2017}]{Johnson2017}
Johnson, J.; Douze, M.; and J{\'e}gou, H.
\newblock 2017.
\newblock Billion-scale similarity search with gpus.
\newblock {\em arXiv preprint arXiv:1702.08734}.

\bibitem[\protect\citeauthoryear{Khandelwal \bgroup et al\mbox.\egroup
  }{2020}]{Khandelwal2020}
Khandelwal, U.; Levy, O.; Jurafsky, D.; Zettlemoyer, L.; and Lewis, M.
\newblock 2020.
\newblock Generalization through memorization: Nearest neighbor language
  models.
\newblock In {\em Proceedings of the 2020 International Conference on Learning
  Representations}.

\bibitem[\protect\citeauthoryear{Lee, Chang, and Toutanova}{2019}]{Kenton2019}
Lee, K.; Chang, M.; and Toutanova, K.
\newblock 2019.
\newblock Latent retrieval for weakly supervised open domain question
  answering.
\newblock {\em CoRR} abs/1906.00300.

\bibitem[\protect\citeauthoryear{Li \bgroup et al\mbox.\egroup
  }{2016}]{Li-etal-2016-diversity}
Li, J.; Galley, M.; Brockett, C.; Gao, J.; and Dolan, B.
\newblock 2016.
\newblock A diversity-promoting objective function for neural conversation
  models.
\newblock In {\em Proceedings of the 2016 Conference of the North {A}merican
  Chapter of the Association for Computational Linguistics: Human Language
  Technologies},  110--119.
\newblock San Diego, California: Association for Computational Linguistics.

\bibitem[\protect\citeauthoryear{Serban \bgroup et al\mbox.\egroup
  }{2016}]{AAAI1611957}
Serban, I.; Sordoni, A.; Bengio, Y.; Courville, A.; and Pineau, J.
\newblock 2016.
\newblock Building end-to-end dialogue systems using generative hierarchical
  neural network models.
\newblock In {\em Proceedings of the Thirtieth AAAI Conference on Artificial
  Intelligence (AAAI 2016)},  3776–--3784.

\bibitem[\protect\citeauthoryear{Serban \bgroup et al\mbox.\egroup
  }{2017}]{serban2017b}
Serban, I.~V.; Klinger, T.; Tesauro, G.; Talamadupula, K.; Zhou, B.; Bengio,
  Y.; and Courville, A.
\newblock 2017.
\newblock Multiresolution recurrent neural networks: An application to dialogue
  response generation.
\newblock In {\em Proceedings of the Thirty-First AAAI Conference on Artificial
  Intelligence (AAAI 2017)}.

\bibitem[\protect\citeauthoryear{Sutskever, Vinyals, and
  Le}{2014}]{sutskever2014sequence}
Sutskever, I.; Vinyals, O.; and Le, Q.
\newblock 2014.
\newblock Sequence to sequence learning with neural networks.
\newblock In {\em Proceedings of Advances in Neural Information Processing
  Systems (NIPS)},  3104--–3112.

\bibitem[\protect\citeauthoryear{Vinyals and Le}{2015}]{Vinyals2015}
Vinyals, O., and Le, Q.
\newblock 2015.
\newblock A neural conversational model.
\newblock In {\em Proceedings of ICML Deep Learning Workshop}.

\bibitem[\protect\citeauthoryear{Wolf \bgroup et al\mbox.\egroup
  }{2019}]{Wolf2019}
Wolf, T.; Sanh, V.; Chaumond, J.; and Delangue, C.
\newblock 2019.
\newblock Transfertransfo: {A} transfer learning approach for neural network
  based conversational agents.
\newblock {\em CoRR} abs/1901.08149.

\bibitem[\protect\citeauthoryear{Xing \bgroup et al\mbox.\egroup
  }{2018}]{AAAIXing}
Xing, C.; Wu, Y.; Zhou, M.; Huang, Y.; and Ma, W.-Y.
\newblock 2018.
\newblock Hierarchical recurrent attention network for response generation.
\newblock In {\em Proceedings of the The Thirty-Second AAAI Conference on
  Artificial Intelligence (AAAI 2018)},  5610–--5617.

\bibitem[\protect\citeauthoryear{Yang \bgroup et al\mbox.\egroup
  }{2019}]{Yang2019}
Yang, Z.; Dai, Z.; Yang, Y.; Carbonell, J.~G.; Salakhutdinov, R.; and Le, Q.~V.
\newblock 2019.
\newblock Xlnet: Generalized autoregressive pretraining for language
  understanding.
\newblock {\em CoRR} abs/1906.08237.

\bibitem[\protect\citeauthoryear{Zhang \bgroup et al\mbox.\egroup
  }{2019}]{Zhang2019}
Zhang, Z.; Han, X.; Liu, Z.; Jiang, X.; Sun, M.; and Liu, Q.
\newblock 2019.
\newblock {ERNIE:} enhanced language representation with informative entities.
\newblock {\em CoRR} abs/1905.07129.

\end{thebibliography}

\end{document}